\documentclass[sigconf]{acmart}
\copyrightyear{2024}
\acmYear{2024}
\setcopyright{acmlicensed}
\acmConference[KDD '24] {Proceedings of the 30th ACM SIGKDD Conference on Knowledge Discovery and Data Mining }{August 25--29, 2024}{Barcelona, Spain.}
\acmBooktitle{Proceedings of the 30th ACM SIGKDD Conference on Knowledge Discovery and Data Mining (KDD '24), August 25--29, 2024, Barcelona, Spain}
\acmISBN{979-8-4007-0490-1/24/08}
\acmDOI{10.1145/XXXXXX.XXXXXX}

\usepackage{multirow}
\usepackage{multicol}
\usepackage{bm}
\usepackage{enumitem}

\usepackage{bbm}
\usepackage{color}
\usepackage{microtype}
\usepackage{booktabs}

\usepackage{makecell}
\usepackage{graphicx}
\usepackage{subcaption}
\usepackage{stfloats}

\AtBeginDocument{%
  \providecommand\BibTeX{{%
    \normalfont B\kern-0.5em{\scshape i\kern-0.25em b}\kern-0.8em\TeX}}}

\acmISBN{978-1-4503-XXXX-X/18/06}

\begin{document}

\title{Path-Specific Causal Reasoning for Fairness-aware Cognitive Diagnosis}

\author{Dacao Zhang}
\email{zhdacao@gmail.com}
\affiliation{
  \institution{School of Computer Science and Information Engineering, Hefei University of Technology}
  \city{Hefei}
  \state{Anhui}
  \country{China}
}

\author{Kun Zhang}
\email{zhang1028kun@gmail.com}
\authornote{Corresponding authors.}
\affiliation{%
  \institution{School of Computer Science and Information Engineering, Hefei University of Technology}
  \city{Hefei}
  \state{Anhui}
  \country{China}
}

\author{Le Wu}
\email{lewu.ustc@gmail.com}
\affiliation{%
  \institution{School of Computer Science and Information Engineering, Hefei University of Technology}
  \institution{Institute of Dataspace, Hefei Comprehensive National Science Center}
  \city{Hefei}
  \state{Anhui}
  \country{China}
}

\author{Mi Tian}
\email{tianmixlb@gmail.com}
\affiliation{%
 \institution{School of Computer Science and Information Engineering, Hefei University of Technology}
 \city{Hefei}
 \state{Anhui}
 \country{China}
}

\author{Richang Hong}
\email{hongrc.hfut@gmail.com}
\affiliation{%
  \institution{School of Computer Science and Information Engineering, Hefei University of Technology}
  \institution{Institute of Dataspace, Hefei Comprehensive National Science Center}
  \city{Hefei}
  \state{Anhui}
  \country{China}
}

\author{Meng Wang}
\email{eric.mengwang@gmail.com}
\affiliation{%
  \institution{School of Computer Science and Information Engineering, Hefei University of Technology}
  \city{Hefei}
  \state{Anhui}
  \country{China}
}

%
\renewcommand{\shortauthors}{Dacao Zhang, Kun Zhang, Le Wu, Mi Tian, Richang Hong, Meng Wang.}

\newcommand{\name}{\emph{PSCRF}}
\newcommand{\fname}{\emph{Path-Specific Causal Reasoning Framework}}
\newcommand{\red}[1]{\textbf{\textcolor{red}{#1}}}

\begin{abstract}
	Cognitive Diagnosis~(CD), which leverages students and exercise data to predict students' proficiency levels on different knowledge concepts, is one of fundamental components in Intelligent Education.
	Due to the scarcity of student-exercise interaction data, most existing methods focus on making the best use of available data, such as exercise content and student information~(e.g., educational context). 
	Despite the great progress, the abuse of student sensitive information has not been paid enough attention. 
	Due to the important position of CD in Intelligent Education, employing sensitive information when making diagnosis predictions will cause serious social issues. 
	Moreover, data-driven neural networks are easily misled by the shortcut between input data and output prediction, exacerbating this problem.  
	Therefore, it is crucial to eliminate the negative impact of sensitive information in CD models. 
	In response, we argue that sensitive attributes of students can also provide useful information, and only the shortcuts directly related to the sensitive information should be eliminated from the diagnosis process. 
	Thus, we employ causal reasoning and design a novel \fname~(\name) to achieve this goal. 
	Specifically, we first leverage an encoder to extract features and generate embeddings for general information and sensitive information of students.  
	Then, we design a novel attribute-oriented predictor to decouple the sensitive attributes, in which fairness-related sensitive features will be eliminated and other useful information will be retained. 
	Finally, we designed a multi-factor constraint to ensure the performance of fairness and diagnosis performance simultaneously.
	Extensive experiments over real-world datasets~(e.g., PISA dataset) demonstrate the effectiveness of our proposed \name. 
\end{abstract}

\ccsdesc[500]{Information systems~Personalization}

\keywords{Cognitive Diagnosis, User modeling, Causal Reasoning, Sensitive Attribute, Fairness}

\maketitle

\section{Introduction}
\label{s:introduction}
As a fundamental component in Intelligent Education, Cognitive Diagnosis~(CD) requires an agent to mine student behavior data to access and identify the student's proficiency level in knowledge concepts~\cite{wang2020neural}. 
It has been applied in various education scenarios, such as student performance prediction~\cite{gao2022deep,yao2021stimuli}, computerized adaptive testing~\cite{wang2023gmocat}, and exercise recommendation~\cite{yin2020mooc,yang2023cognitive}. 

\begin{figure}[htbp]
	\vspace{-4mm}
	\begin{center}
		\includegraphics[width=0.45\textwidth]{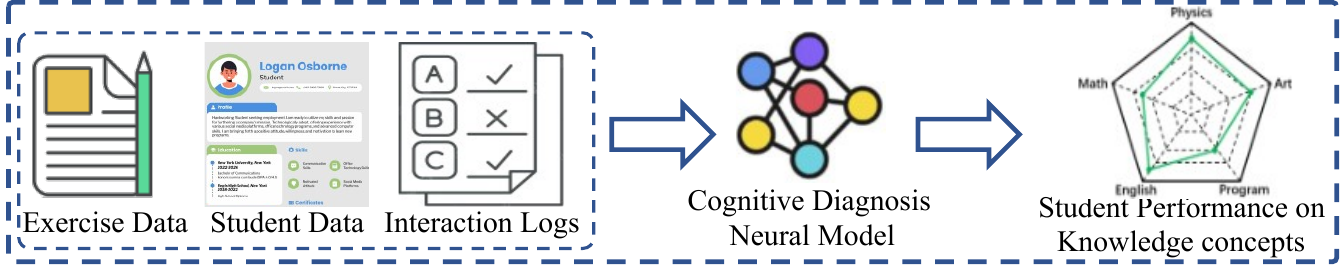}
	\end{center}
	\vspace{-3mm}
	\caption{A tiny example of general cognitive diagnosis.  }
	\label{f:example}
	\vspace{-4mm}
\end{figure}

Literately, researchers have designed enormous neural network-based methods to realize accurate diagnosis and student modeling. 
As illustrated in Figure~\ref{f:example}, existing models usually take multi-type information~(e.g., student-exercise interaction logs, student's personal information, exercise text) as the input, and predict students' mastery of each concept. 
For example, \citet{wang2020neural} have built a deep full connection neural network to capture the complex student-exercise interaction records. 
\citet{wang2022neuralcd} designed a novel KaNCD method to address the weak knowledge concepts coverage problem. 
Besides, there are many other methods for high-order student-exercise interaction modeling~\cite{zhou2021modeling,bao2021exploring,cheng2019enhancing,liu2021towards} and graph-based modeling~\cite{gao2021rcd,li2022hiercdf,yu2024rdgt,liu2023homogeneous}.

\begin{table}[t]
	\centering
	\small
	\caption{The probability of students answering questions correctly (i.e., Data Statistics) and the predicted probability of students answering questions correctly by NCD and KaNCD among different groups. The groups are divided by the sensitive attributes of students~(i.e., family wealth or country).}
	\vspace{-4mm}
	\begin{tabular}{l|ccc|cc}
		\toprule
		\multirow{2}{*}{Model} & \multicolumn{3}{c|}{Family Wealth} & \multicolumn{2}{c}{Country} \\ \cline{2-6}
		& Poor      & Average    & Wealth   & Australia      & Brazil     \\
		\midrule
		Data statistics  & 0.4736    & 0.5448     & 0.6434   & 0.5516   & 0.3888  \\
		NCD              & 0.5140    & 0.5861     & 0.6789   & 0.5913   & 0.3293  \\
		KaNCD            & 0.4778    & 0.5589     & 0.6643   & 0.5650   & 0.3025  \\ 
		NCD-\name~ & 0.5545 & 0.5798 & 0.6155 & 0.5824 & 0.3321 \\
		KaNCD-\name~ & 0.5286 & 0.5581 & 0.6271 & 0.5680 & 0.3026 \\
		\bottomrule      
	\end{tabular}
	\label{t:biased-results}
	\vspace{-6mm}
\end{table}
Despite the great progress, these methods will inevitably introduce fairness issues while exploiting the full potential of student data. 
Taking Table~\ref{t:biased-results} as an example, by conducting statistical analysis, we can observe that students who are from more affluent families or in more developed areas usually have better performance on exercise (e.g., $0.6434$ points for rich boys v.s. $0.4736$ points for poor boys). 
In fact, this phenomenon occurs more because those students receive better support or training~(e.g., more books, computer access opportunities, etc), rather than better family circumstances. 
However, if we do not constrain the model to exploit all the data, it will easily learn the connections between sensitive information of students and student performance (e.g., using family wealth to predict the student proficiency level), which we name as spurious correlations. 
From the results in Table~\ref{t:biased-results}, we can observe this type of phenomenon. 
By using PISA data~\cite{oecd2015} to train NCD and KaNCD models directly, they will overestimate these advantaged students (e.g., $0.565$ for Australia v.s. $0.3025$ for Brazil), showing that they have taken advantage of sensitive information and made unfair predictions. 
If we apply the unfair results to real-world scenarios, it will exacerbate social prejudices and conflicts, bringing about bad social effects. 
More seriously, according to the results in Table~\ref{t:biased-results}, even if we do not use sensitive attributes as model inputs, NCD and KaNCD models still can infer sensitive attributes of students from the interaction logs and abuse them for better performance.
Therefore, \textbf{it is crucial to exclude the abuse of student sensitive attributes while ensuring the diagnosis performance}. 

Recently, plenty of fairness-aware methods have been proposed, such as data reweighting (resampling)~\cite{iosifidis2019fae,rastegarpanah2019fighting} and adversarial learning~\cite{bose2019compositional,wu2021learning,zhang2024understanding}. 
However, these strategies still have unavoidable shortcomings. 
For example, data resampling methods usually increase/decrease weights of certain student-exercise interactions to realize the fairness target. 
However, this strategy violates the principle in cognitive diagnosis that the same student should only respond to the same exercise once, and is also dependent on the sensitive attributes~\cite{james2023resampling}. 
Meanwhile, adversarial learning uses an additional classifier to predict the sensitive attribute from user embeddings and eliminate corresponding information directly. 
This strategy is too coarse-grained to distinguish available information from sensitive information, leading to a decrease in model capability. 
To answer the above question, we propose that the fairness-related sensitive features from sensitive attributes should be eliminated as comprehensively as possible while diagnosis-related features from sensitive attributes should be retained as much as possible.
For example, family wealth cannot be used as an influencing factor in determining the student proficiency level, while the quality of the learning environment can. 
For this goal, causal inference~\cite{pearl2016causal} is one promising direction. 
By distinguishing causation and correlations from biased real-world data, causal inference has made great progress in  medicine~\citep{Konigorski2021CausalII}, neuroscience~\citep{Marinescu2018QuasiexperimentalCI}, cognitive science~\citep{Shams2010CausalII}, etc. 
It also has been proven useful in addressing bias issues in vision question answering~\cite{niu2021counterfactual}, text classification tasks~\cite{qian2021counterfactual}, anomaly detection~\cite{wu2022bias}, and so on.

To this end, in this paper, we propose to employ causal inference and design a novel \fname~(\name) for fairness-aware CD modeling. 
Specifically, we leverage a causal graph to describe the correlations and causation between different factors and student proficiency levels.  
Based on the causal graph, we try to use \name~to calculate the path-specific effect of different inputs to the output. 
We first leverage an encoder to extract features from student-exercise interaction logs and generate embeddings for student IDs and sensitive attributes. 
Next, we design a novel attribute-oriented predictor (Decoupled Predictor~(DP)) to realize the decoupling of sensitive attributes and useful information, in which fairness-related sensitive feature embeddings are used to predict the sensitive attributes and diagnosis-related feature embeddings are used to predict the useful information from sensitive attributes.
Moreover, to ensure the quality of decoupling, we also design a multi-factor fairness constraint to restrict the distance of different embeddings. 
Then, the fairness-aware inference can be obtained by removing the fairness-related sensitive features from the diagnosis process. 
Finally, we conducted extensive experiments over real-world diagnosis data in various settings. Experimental results demonstrate that \name~can achieve impressive debiased performance while maintaining the accuracy of student proficiency level modeling. 
We also release the code to facilitate the community\footnote{https://github.com/NLPfreshman0/PSCRF}.

In summary, our main contributions are reported as follows:
\begin{itemize}
	\item We argue that student sensitive information can provide useful information for diagnosis performance improvement. Thus, we should eliminate the fairness-related spurious correlations and retain useful information simultaneously for better fairness-aware CD. 	
	\item We introduce causal inference into the diagnosis process and design a novel \name~to realize the fairness-aware CD, in which path-specific causal reasoning is employed to eliminate the abuse of sensitive attributes. 
	
	\item We have conducted extensive experiments over real-world data and provided detailed analyses of results and models to demonstrate the effectiveness of \name.
\end{itemize}

\section{Related Work}
\label{s:related-work}

The related work can be summarized into two components:
1) \textit{Cognitive Diagnosis}: giving a brief introduction of CD in intelligent education scenarios;
2) \textit{Fairness-aware User Modeling}: focusing on fair user representation learning from biased real-world data.

\subsection{Cognitive Diagnosis}
Cognitive Diagnosis (CD) is a fundamental and pivotal task in many real-world intelligent education scenarios~\cite{daud2017predicting,van2000computerized}. 
It requires an agent to predict students' proficiency level of each knowledge concept through historical student-exercise interaction logs.  
DINA~\cite{dina2009dina} and IRT~\cite{embretson2013item} are two representative methods in this domain. 
DINA is a discrete CDM that assumes student mastery levels are binary (master the knowledge concept or not)~\cite{dina2009dina}.  
IRT characterizes students' abilities as unidimensional and continuous latent traits and designs logistic-like interaction functions to model the probability of a student correctly answering an exercise~\cite{embretson2013item}. 

To improve the diagnosis performance, various methods have been proposed to extend the capability of DINA and IRT~\cite{gao2023leveraging,tong2021item,huang2021group,wang2023self} and exploit the potential of student and exercise data. 
For example, \citet{cheng2019enhancing} proposed a DIRT method to extract semantic features from the content of exercise texts for high-quality representation generation. 
\citet{wang2020neural} designed an NCD method to exploit student-exercise interactions for accurate student proficiency level modeling. 
Moreover, \citet{zhou2021modeling} proposed to improve CD performance from the student perspective. 
They employed context and culture information of students to enrich the student proficiency representation, which is in favor of improving the diagnosis performance. 
Besides, other data issues in CD are also considered, such as weak knowledge concepts coverage problems~\cite{wang2022neuralcd} and non-interactive knowledge concepts problems~\cite{ma2022knowledge}.

\subsection{Fairness-aware User Modeling}
User modeling focuses on measuring user characteristics based on user-related data, which plays a crucial role in plenty of scenarios, such as user preference modeling in recommender system~\cite{wu2022survey} and user proficiency level modeling in Intelligent Education~\cite{shen2022assessing}. 
Recent studies have demonstrated that user-related data may contain stereotypes or biased data, which will mislead models to learn the spurious correlations and make vulnerable and unfair decisions. 
To alleviate this problem, plenty of fairness-aware user modeling methods have been proposed, such as data reweighting~\cite{iosifidis2019fae,rastegarpanah2019fighting,geyik2019fairness}, regularization~\cite{cheng2019enhancing,yao2017beyond}, and adversarial learning~\cite{wu2021learning,zhu2020measuring,beigi2020privacy,zhang2024understanding}.
Among all these methods, causal inference-based methods are one promising direction. 
For example, 
\citet{zhao2022popularity} proposed a disentangled framework TIDE based on path-specific causal reasoning to deal with the popularity bias in user preference modeling in recommendations. 
\citet{chen2023data} designed a novel data augmentation strategy to balance the training data, so that sensitive-related information will be inactivated when modeling user preference. 
Apart from this, other types of biases are also hot research topics, such as selection bias~\cite{marlin2009collaborative,chen2023bias},  exposure bias~\cite{chen2018modeling}, and unfairness~\cite{ekstrand2018exploring,chen2023data}. 

However, due to the sparsity characteristic of student and exercise data in education, existing methods mainly focus on exploiting the potential of data, ignoring the implicit sensitive information abuse problem. 
Since education plays a crucial role in influencing the trajectory of individuals' adult lives~\cite{simon2007education}, it is urgent to focus more on this problem. 
Some works have made early attempts. 
For example, \citet{yu2020towards}~conducted an analysis to explore the equitable prediction of short-term and long-term college success using various sources of student data.
\citet{li2021user}~proposed a Fair-LR algorithm to achieve accurate and fair AI prediction to help to realize fair student modeling. 
\citet{zhang2024understanding}~divided student performance into bias proficiency and fair
proficiency, then used only fair
proficiency to make predictions.
For fairness-aware CD modeling, enormous works remain unexplored, such as fair student representations, sensitive attribute utilization, and so on.

\textbf{Our Distinction.}
We focus on a more impactful issue: \textit{How to eliminate the abuse of student sensitive attributes from CD models while ensuring the diagnosis performance?} 
We argue that student sensitive attributes can also provide useful information, so directly removing them from CD models is not optimal. 
Thus, we design a novel \name~to realize the debiased CD learning while retaining the diagnosis performance. 
Specifically, \name~decouples student sensitive attributes into sensitive-related information that should not be used in diagnosis process, and sensitive-unrelated information that can be used to improve the diagnosis performance. 
Moreover, \name~leverages a multi-factor normalization to ensure the quality of debiased learning and diagnosis performance simultaneously.

\section{Preliminary}
\label{s:preliminary}

\subsection{Prerequisite Knowledge}
\label{s:prerequist}
\begin{figure}[htbp]
	\vspace{-4mm}
	\begin{center}
		\includegraphics[width=0.4\textwidth]{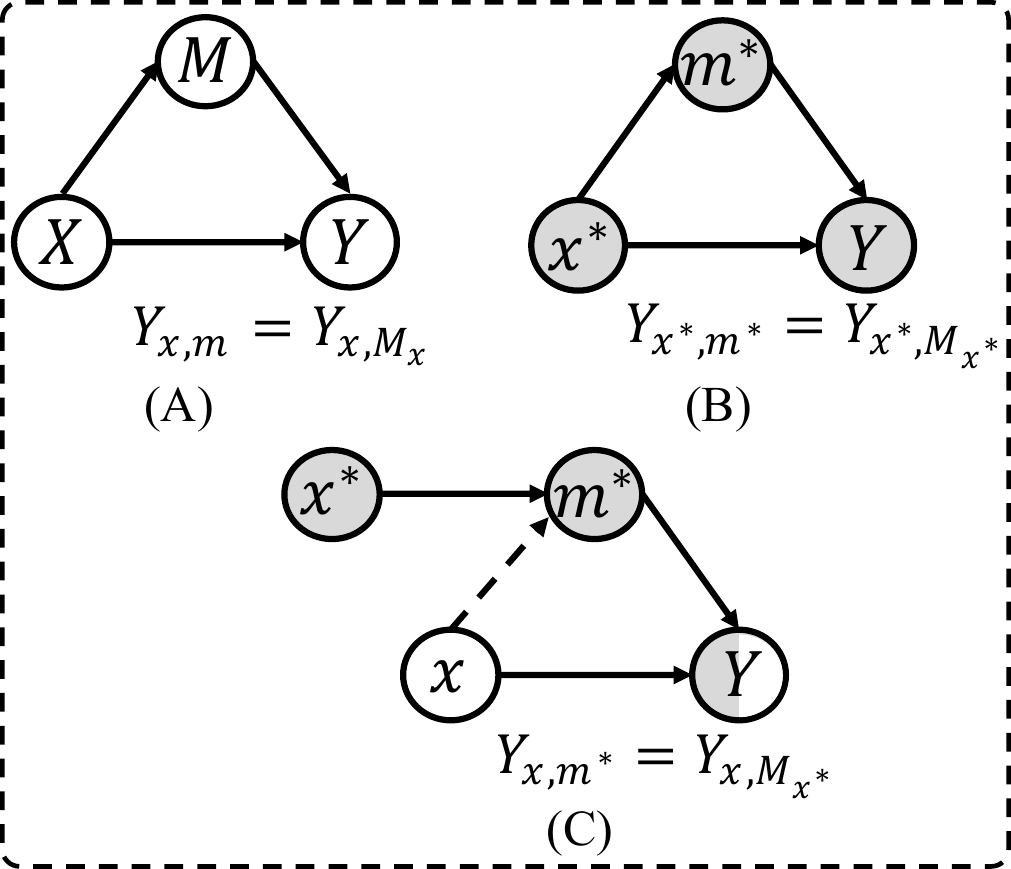}
	\end{center}
	\vspace{-4mm}
	\caption{The general causal graph and commonly used counterfactual notations.}
	\label{f:basic_cgraph}
	\vspace{-4mm}
\end{figure}
This section explains the causal graph and causal effect calculation mentioned, aiming to help readers understand the significance and importance of the causal graph.

\textbf{Causal Graph. }
The causal graph is a Directed Acyclic Graph (DAG) $\mathcal{G}=<\mathcal{V}, \mathcal{E}>$, which describes the causal relationships between different variables. 
$\mathcal{V}$ is the node set and $\mathcal{E}$ is the edge set. 
As illustrated in Figure~\ref{f:basic_cgraph}(A), the arrow indicates the direction of causality. 
For example, $X \rightarrow Y$ denotes that variable $X$ has a direct effect on $Y$. $X \rightarrow M \rightarrow Y$ denotes that variable $X$ has indirect effect on $Y$ through mediator $M$. Following these notations, assume $X=x$, then the value of $Y$ can be calculated as follows: 
\begin{equation}
	\label{eq:causal-basic}
	Y_{x,m} = Y(X=x, M=m=M_x), \\
\end{equation}
where the value of mediator $M$ can be calculated with $m=M_x=M(X=x)$. 
\begin{figure}[htbp]
	\begin{center}
		\includegraphics[width=0.45\textwidth]{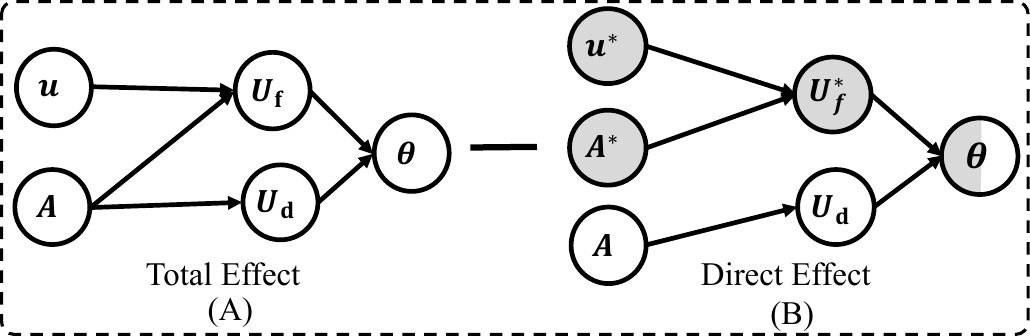}
	\end{center}
	\vspace{-4mm}
	\caption{The causal graph used in our proposed \name.}
	\label{f:cause_graph}
	\vspace{-4mm}
\end{figure}

\textbf{Causal Effect Calculation.}
The causal effect is a comparison of the potential outcomes of giving two different interventions to the same variable. 
As shown in Figure~\ref{f:basic_cgraph}(A)-(B), assume that $X=x$ is the treatment, then $Y_{x,m}$ denotes the potential outcome of the treatment. Similarly, $X=x^*$ is the no treatment, then $Y_{x^*,m^*}$ is the potential outcome of no treatment. 
Along this line, the causal effect can be calculated as follows:
\begin{equation}
	\label{eq:causal-cal}
	Effect = Y_{x,m} - Y_{x^*, m^*} = P(y|\textit{do}(X=x)) - P(y|\textit{do}(X=x^*)), \\
\end{equation}
where $\textit{do}(X=x)$ and $\textit{do}(X=x^*)$ is the intervention to the variable $X$. Note that $\textit{do}(\cdot)$ operation requires that only the treatment variable is intervened, all other variables are not intervened. 
To satisfy this requirement, the counterfactual operation is proposed.

\textbf{Counterfactual Notation. }
The counterfactual reason is to assume a scenario that all the other variables remain unchanged and only the treatment variable is changed. 
The causal effect of the treatment variable on the output variable can be calculated in this scenario. 
For example, in Figure~\ref{f:basic_cgraph}(C), $Y_{x, M_{x^*}} = Y(X=x, M=M(X=x^*))$ denotes a typical counterfactual reasoning.

\subsection{PISA Data Introduction}
As mentioned above, the educational context of students can be used to obtain better student representations, which refer to the various features related to students' learning process~\cite{zhou2021modeling}. 
OECD’s Programme for International Student Assessment~(PISA) focused on this topic and designed multiple questions to investigate and collect these educational contexts, such as family wealth, education degree of parents, and so on. 
For example, when investigating the highest education degree of students' parents, five options~(e.g., 1-General senior, 2-Vocational senior, 3-Junior) are provided. 
Students can select one option based on their situations. 
This information can be used as student attributes. 
Table~\ref{t:context-example} lists some examples of these questions. 
Moreover, this organization has developed exercises to measure 15-year-olds' ability to employ reading, mathematics, and science knowledge and skills to meet real-life challenges~\cite{oecd2015}. 
They investigated students from different countries, and released the data and technical reports on a three-year cycle, which is suitable for student attribute-aware cognitive diagnosis. 
\begin{table}[htbp]
	\vspace{-4mm}
	\small
	\centering
	\caption{Educational Context examples from PISA dataset}
	\vspace{-4mm}
	\begin{tabular}{c|l}
		\toprule
		Aspect                  & Question Examples                                          \\
		\midrule
		\multirow{3}{*}{Home}   & Home Economic, Social and Cultural Status (ESCS)           \\\cline{2-2}
		& Highest education degree of parents                        \\\cline{2-2}
		& Number of equipment, appliances, and rooms                 \\
		\hline
		\multirow{3}{*}{Person} & Whether students have a grade repetition experience        \\\cline{2-2}
		& how many days did students engage in out-school activities \\
		\bottomrule
	\end{tabular}
	\label{t:context-example}
	\vspace{-4mm}
\end{table}

With the guidance of technical report~\cite{oecd2015}, we select ESCS index as the sensitive attribute example to tackle the problem in Section~\ref{s:introduction}. Moreover, we leverage the Pearson Correlation Coefficient to select the useful but not sensitive attributes: \textit{1) The number of books, 2) The number of tablet computers, 3) A link to the Internet, 4) A computer can be used for school work, 5) The number of E-book readers.}
These selected attributes all exhibit strong correlations with ESCS index, which we have reported the results in Table~\ref{tab:useful1} and Table~\ref{tab:useful2} in the Appendix. 
However, they are not sensitive attributes and directly contribute to the development of students' abilities, which should be considered in the diagnosis process.

\begin{table}[htbp]
	\vspace{-4mm}
	\small
	\centering
	\caption{Notations and explanations in our proposed \name.}
	\vspace{-4mm}
	\begin{tabular}{c|l}
		\toprule
		Notation                  & Explanation \\
		\midrule
		$\bm{u}, \bm{u}*$ & Student ID embeddings and their counterfactuals \\
		$A, A^*$ & sensitive attribute and counterfactual sensitive attribute\\
		$\bm{U}_f, \bm{U}_f^*$ & The diagnosis-related features and their counterfactuals\\
		$\bm{U}_d$ & The fairness-related feature \\
		$\alpha$ & Learnable parameters for integrating $\bm{U}_f$ and $\bm{U}_d$ \\
		$\phi_e$ & Parameters related to exercises(e.g., difficulty, discrimination) \\
		$\bm{\theta}, \bm{\theta}^*$ & The student proficiency level and their counterfactuals \\
		$\beta$ & Learnable parameters controlling the degree of debiasing \\
		$\bm{\theta}_d$ & The fairness-aware student proficiency level \\
		\bottomrule
	\end{tabular}
	\label{t:notation}
	\vspace{-6mm}
\end{table}
\section{Technical Details of \name}
\label{s:model}

\subsection{Causal view of \name}
\label{s:causal-view}
Based on the motivation in Section~\ref{s:introduction}, we use the causal graph in Figure~\ref{f:cause_graph}(A) to describe the causal relation among different paths.
Specifically, $\bm{u}$ denotes the general representation of one student (i.e., ID embeddings). 
$\bm{A}$ is the corresponding sensitive attribute representation (e.g., ESCS embeddings). 
We argue that sensitive attributes contain fairness-related sensitive features and diagnosis-related features. 
The former should not be used in the diagnosis process while the latter should be used to improve the diagnosis performance. 
Therefore, we leverage the causal path $\bm{A} \rightarrow \bm{U}_d \rightarrow \bm{\theta}$ to denote the effect of fairness-related sensitive features, which should be removed from the entire graph. 
A toy example is that family wealth should not be considered in the diagnosis process since it will introduce unfairness to vulnerable groups. 

Meanwhile, we use the causal path $(\bm{u}, \bm{A}) \rightarrow \bm{U}_f \rightarrow \bm{\theta}$ to denote the effect of diagnosis-related features from sensitive attributes and general information. 
one similar example is as follows: though family wealth cannot be used in the diagnosis process, we can exploit \textit{the number of books} or \textit{a link to the Internet} to better model student proficiency level since they directly contribute to the student's ability development. 
Based on this causal graph, we then introduce the corresponding implementation.

According to the principle of Average Treatment Effect~(ATE), we can calculate the Total Effect (TE) of all input variables to the output prediction as follows:
\begin{equation}
	TE = \bm{\theta}(u, A) -\bm{\theta}(u^*, A^*).
\end{equation}

Next, we intend to calculate the effect of only fairness-aware sensitive features on the output prediction. For this target, we employ the Natural Direct Effect~(NDE) as follows:
\begin{equation}
	NDE = \bm{\theta}(u^*, A) -\bm{\theta}(u^*, A^*).
\end{equation}

Finally, we can obtain the debiased prediction by calculating the Total Indirect Effect (TIE) as follows:
\begin{equation}
	TIE = TE - NDE = \bm{\theta}(u, A) - \bm{\theta}(u^*, A).
\end{equation}
By maximizing the Total Indirect Effect (TIE) inference, we can achieve debiased learning in the diagnosis process.

\begin{figure*}[htbp]
	\begin{center}
		\includegraphics[width=0.9\textwidth]{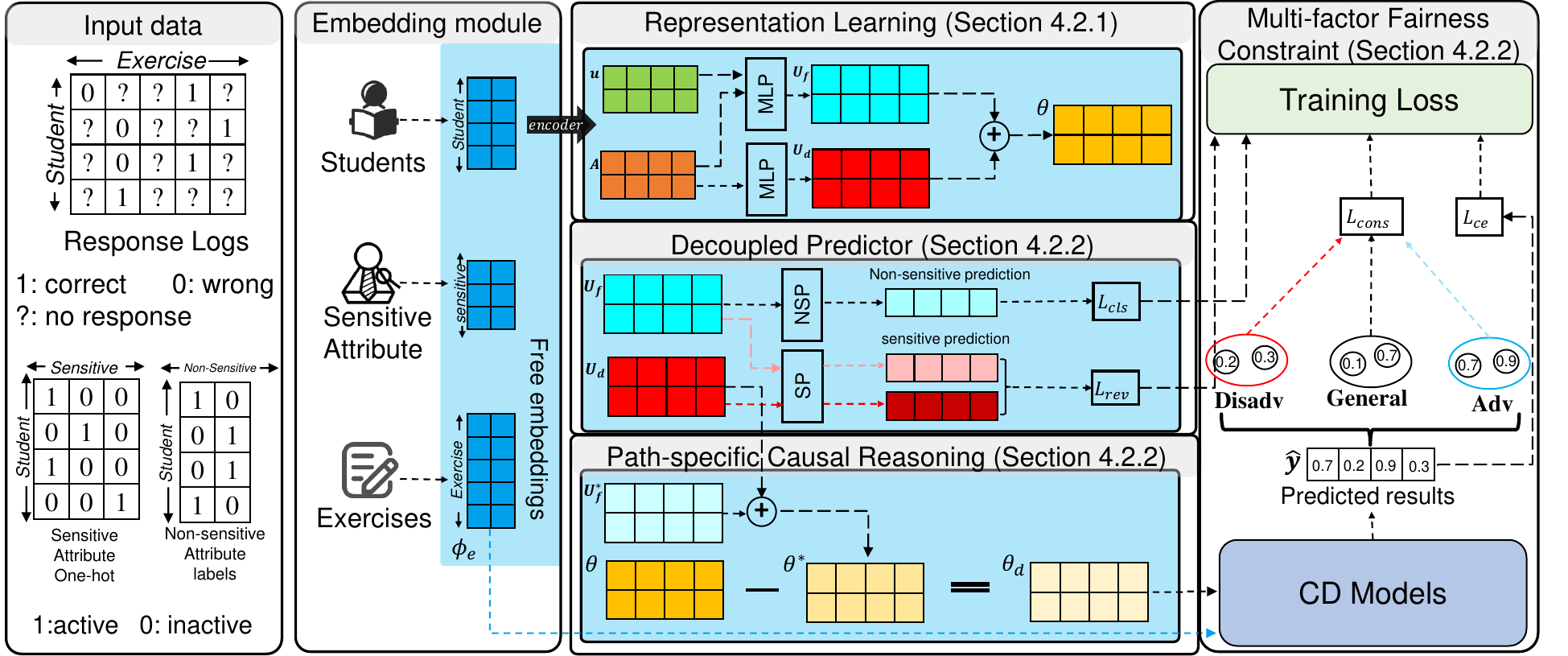}
	\end{center}
	\vspace{-4mm}
	\caption{The overall framework of our proposed \name~}
	\label{f:model}
	\vspace{-4mm}
\end{figure*}

\subsection{Causal Implementation: \name}
Figure~\ref{f:model} illustrates the overall architecture of \name, which consists of two main components: \textit{Representation Learning Module} and \textit{Decoupled and Constraint Module}. Next, we introduce each component in detail. Table~\ref{t:notation} explains the notations in \name.

\subsubsection{Representation Learning Module}
Similar to previous works~\cite{wang2020neural,zhou2021modeling}, we first embed input entities into latent embeddings, which including student embeddings $\bm{U} = [\bm{u}_1, \bm{u}_2, ..., \bm{u}_N]^\top \in \mathbb{R}^{N\times d}$, exercise embeddings $\bm{V} = [\bm{v}_1, \bm{v}_2, ..., \bm{v}_M]^\top \in \mathbb{R}^{M\times d}$, and sensitive attribute embeddings $\bm{A} = [\bm{a}_1, \bm{a}_2, ..., \bm{a}_T]^\top \in \mathbb{R}^{T\times d}$.
$d$ is the embedding dimension. 
These embeddings are randomly initialized and will be updated during model learning. 

After obtaining the embeddings of different entities, we intend to learn the representations of fairness-related sensitive features $\bm{U}_d$ and diagnosis-related features $\bm{U}_f$, which is realized by our designed proficiency modeling module. 
Specifically, we employ a fairness-related sensitive feature generator to obtain $\bm{U}_d$, which takes sensitive attribute embeddings as input and uses a Multi-Layer Perceptron~(MLP) to generate representations as follows: 
\begin{equation}
	\label{eq:instein}
	\bm{U}_d^i = \sigma(\textit{MLP}_1(\bm{A}_{[i]})), \\
\end{equation}
where $\sigma(\cdot)$ represents the sigmoid function, ${\small \bm{A}_{[i]}}$ is the sensitive attribute embedding set for the $i^{th}$ student. $\bm{U}_d^i$ is the sensitive-related feature representation of the $i^{th}$ student.

Similarly, a diagnosis-related feature extractor is used to generate $\bm{U}_f$ with student ID embeddings and sensitive attribute embeddings. Another MLP is used to obtain the representations:
\begin{equation}
	\label{eq:uninstein}
	\bm{U}_f^i = \sigma(\textit{MLP}_2(\textit{concat}(\bm{u}_i, \bm{U}_d^i))), \\
\end{equation}
where  $\textit{concat}(\cdot)$ denotes the concatenation operation. $\bm{u}_i$ is the free ID embedding of the $i^{th}$ student. 

After obtaining the results from two extractors, we leverage a learnable parameter $\alpha$ to fuse these two different features and generate student proficiency level $\bm{\theta}_i$ as:
\begin{equation}
	\bm{\theta}_i = \sigma((1 - \alpha) \bm{U}_f^i + \alpha \bm{U}_d^i).\\
	\label{eq:fusion}
\end{equation}

\subsubsection{Decoupled and Constraint Module}
\label{s:constraint}
In the above module, we aim to extract fairness-related sensitive features and diagnosis-related features from input data. 
However, there are no explicit supervised signals. which poses a big challenge. 
In response, we design the following three modules. 
Next, we omit the student index $i$ for simplicity and introduce how we construct each module.

\textbf{(1) Decoupled Predictor~(DP). } First, we intend diagnosis-related feature embedding $\bm{U}_f$ should include useful information from sensitive attributes and maintain general information of students. 
Therefore, we leverage the educational context as the guidance, and identify top-k educational context questions most correlated with the sensitive attribute by computing the Pearson correlation coefficient between them. 
These educational contexts often relate to students' learning environments and do not involve sensitive attributes (e.g., the number of books), which can be used to enhance the student proficiency level modeling. 
Thus, we leverage selected educational contexts as the prediction targets and formulate the optimization target as follows:
\begin{equation}
	\mathcal{L}_\text{cls} = \frac{1}{K} \sum_{k=1}^{K} \textit{CE}(\textit{MLP}(\bm{U}_f), Label_k), \\
	\label{eq:non-sensitive}
\end{equation}
Where $K$ represents the total number of non-sensitive attributes, $Label_k$ represents the label of the $k^\text{th}$ non-sensitive attribute, and CE denotes the cross-entropy function. 

Meanwhile, to ensure $\bm{U}_d$ to focus only on the fairness-related sensitive features, we develop a novel sensitive attribute enhancement module.
We first send $\bm{U}_d$ to an MLP to predict sensitive attributes, so that $\bm{U}_d$ can be better learned to represent sensitive attributes. 
Meanwhile, $\bm{U}_f$ should not contain these sensitive attributes. 
Thus, we send $\bm{U}_f$ to the same MLP to predict the counterfactual sensitive attributes. 
Therefore, $\bm{U}_d$ can only encode fairness-related sensitive features that should not be used in the diagnosis process. 
$\bm{U}_f$ will not include these sensitive features, which is in favor of fairness-aware diagnosis.
This process can be formulated as follows:
\begin{equation}
	\mathcal{L}_{\text{rev}} = \mathcal{L}(\textit{SMLP}(\bm{U}_d), A) + \mathcal{L}(\textit{SMLP}(\bm{U}_f), A^*),
	\label{eq:sensitive}
\end{equation}
where $\mathcal{L}$ represents the loss function, which can be mean squared error (MSE) for continuous values or cross-entropy for discrete values. 
$\textit{SMLP}(\cdot)$ denotes the shared MLP. 
$A^*$ represents the counterfactual sensitive attribute label, which we will give a detailed explanation in the next section.

\textbf{(2) Path-specific Causal Reasoning.} 
According to Figure~\ref{f:cause_graph} and Section~\ref{s:causal-view}, we need to remove the path $\bm{A} \rightarrow \bm{U}_d \rightarrow \bm{\theta}$ to realize the fairness-aware diagnosis. 
Following the principle of Average Treatment Effect~(ATE), we need to imagine a counterfactual world, which is shown in Figure~\ref{f:cause_graph}(B). 
In the counterfactual world, only the path $\bm{A} \rightarrow \bm{U}_d \rightarrow \bm{\theta}$ remains unchanged. 
We need to block the effect of the path $(\bm{u}, \bm{A})\rightarrow \bm{U}_f \rightarrow \bm{\theta}$. 
Thus, we use the counterfactual sensitive attributes to modify Eq.(\ref{eq:uninstein}) and Eq.(\ref{eq:fusion}) as follows:
\begin{equation}
	\label{eq:counterfactual}
	\begin{split}
		\bm{U}_f^{*} &= \sigma(\textit{MLP}(\textit{concat}(\bm{u}^*, \bm{U}_d^{*}))), \\
		\bm{\theta}^* &= \sigma((1 - \alpha) \bm{U}_f^{*} + \alpha \bm{U}_d), \\
	\end{split}
\end{equation}
where $\{\bm{u}^*, \bm{U}_d^{*}\}$ are the counterfactual student representation and counterfactual fairness-related sensitive feature representation. 
For implementation, we use the mean representation of all student representations
to realize $\bm{u}^*$, and use the mean representation of corresponding sensitive attributes to calculate $\bm{U}_d^{*}$. 
According to the causal inference, this intervention can help \name~to calculate the accurate effect of path $\bm{A} \rightarrow \bm{U}_d \rightarrow \bm{\theta}$. 
Then, we can obtain fairness-aware student proficiency level as follows:
\begin{equation}
	\label{eq:debiased-result}
	\bm{\theta}_d = \sigma(\bm{\theta} - \beta\bm{\theta}^*), \\
\end{equation}
where $\beta$ is a learnable parameter to control the degree of debiasing.
$\bm{\theta}_d$ is used to realize the fairness-aware CD modeling.

\textbf{(3) Multi-factor Fairness Constraint.} 
To facilitate better fairness-aware diagnosis, we introduce a Multi-factor Fairness Constraint. 
Specifically, we partition students into disadvantaged, general, and advantaged groups based on the value of sensitive attributes. 
Since $\bm{\theta}_d$ is the fairness-aware student proficiency level representation, we intend the variance of the predicted means for different groups to be as low as possible. 
Meanwhile, since $\bm{U}_d$ should incorporate unwanted sensitive information as much as possible, we maximize the variance of the predicted means for different groups. 
Therefore, this constrain can be realized as follows:

\begin{equation}
	\mathcal{L}_{\text{cons}} = \textit{std}\left(\overline{y}_{\text{dis}}, \overline{y}_{\text{gene}}, \overline{y}_{\text{adv}}\right)_{\bm{\theta}_d} - \text{std}\left(\overline{y}_{\text{dis}}, \overline{y}_{\text{gene}}, \overline{y}_{\text{adv}}\right)_{\bm{U}_d},
	\label{eq:fair}
\end{equation}
where $\overline{y}_{\text{dis}}$, $\overline{y}_{\text{gene}}$ and $\overline{y}_{\text{adv}}$ respectively denote the predicted means of three groups, while $\textit{std}(\cdot)$ represents the variance.

\subsection{Model Training} 
To ensure prediction accuracy, we also add traditional cross-entropy constraints to $\bm{U}_f$, $\bm{U}_d$, $\bm{\theta}$ and $\bm{\theta}_d$. First, we input them into the CD model to obtain prediction results:
\begin{equation}
	Y_{\hat{\bm{\theta}} } = \textit{CDM}(\hat{\bm{\theta}}, \bm{\phi}_e), \\
\end{equation}
where $CDM(\cdot)$ represents the CD models such as NCD or KaNCD. $\hat{\bm{\theta}}$ can be $\bm{U}_f$, $\bm{U}_d$, $\bm{\theta}$ or $\bm{\theta_d}$. $\bm{\phi}_e$ represents the parameters related to exercises(e.g., difficulty, discrimination). Then we minimize the cross-entropy loss between the predictions and labels:
\begin{equation}
	\mathcal{L}_{\text{ce}} = 
	\sum\limits_{\theta_i \in \Theta}\text{CE}(Y_{\theta_i}, y),
	\label{eq:Lce}
\end{equation}
where $\Theta$ = \{$\bm{U}_f$, $\bm{U}_d$, $\bm{\theta}$, $\bm{\theta_d}$\}. $y$ is the true label. 
Finally, total loss is:
\begin{equation}
	\mathcal{L}_{\text{total}} = w_1\mathcal{L}_{\text{ce}} + w_2\mathcal{L}_\text{cls} + w_3\mathcal{L}_{\text{rev}} + w_4\mathcal{L}_{\text{cons}}, \\
	\label{eq:training}
\end{equation}
where $w_1$,$w_2$, $w_3$, and $w_4$ represent hyperparameters that balance the weights of each part of the loss.


\section{Experiments}
\label{s:experiments}

\subsection{Experimental Setup}
\textbf{Data preprocessing.} 
We procured two prototypical datasets from the PISA-2015, representing Australia and Brazil, respectively, meticulously arranged in descending order based on their developmental status and the mean scholastic attainment of students\cite{oecd_result}. In each dataset, there are 28 different self-acquired features \cite{blau1967american}, such as learning interests and self-efficacy. We used two representative sensitive attributes, namely  ESCS (Index of Economic, Social, and Cultural Status) and the father's education level \cite{oecd2015} to evaluate our method. Specifically, based on the data provided by PISA~\cite{oecd2015}, we categorized students into three groups - \textit{disadvantaged, general, and advantaged} - according to their sensitive attributes, for fairness-aware diagnosis. We filtered out students with fewer than 10 exercise records to ensure sufficient data for training. The Basic statistics of datasets are shown in Table~\ref{tab:dataset}. For each dataset, we performed a $70\%/10\%/20\%$ training/validation/testing split. 
\begin{table}[htbp]
	\vspace{-3mm}
	\centering
	\caption{The Statistics of datasets}
	\vspace{-4mm}
	\label{tab:dataset}
	\begin{tabular}{l|c|c|c}
		\toprule
		Dataset & Students & Exercises & Exercise Records \\
		\midrule
		Australia & 8,485 & 184 & 249,727 \\
		Brazil & 5,777 & 183 & 143,314 \\
		\bottomrule
	\end{tabular}
	\vspace{-3mm}
\end{table}

\textbf{Evaluation Metrics. }
Based on the target, we select two types of metrics. 
For diagnosis performance, following previous works~\cite{gao2021rcd,wang2022neuralcd}, we used widely used metrics: Area Under Curve (AUC) and Accuracy (ACC). 
Meanwhile, following the work~\cite{wang2020neural}, we also use the Degree of Agreement (DOA) for validation. 

For fairness performance, since the abuse of sensitive attributes will mislead models to underestimate or overestimate the students from different groups, commonly used fairness metrics are used.
We first employ Equal opportunity (EO)~\citep{hardt2016equality}:
\begin{equation}
	EO=\textit{Std}(TPR_{disadv}, TPR_{gene},TPR_{adv},), \\  
	\label{eq:eq9}
\end{equation}
$TPR$ refers to True Positive Rates. 
$\textit{Std}(\cdot)$ is the standard deviation. 
Since there are only two situations (i.e., correct and incorrect) for students answering questions, CD models should have equal capability of predicting the probability of students answering exercises correctly across different groups. 
Along this line, sensitive attributes can be proved to be not used in the diagnosis process. 
Moreover, since predictions of CD models have a big social influence in real-world scenarios, we argue that the rights of vulnerable groups should be guaranteed. 
We should not be prejudiced against disadvantaged groups and assume that they will perform less well. 
\begin{table*}[t]
	\centering
	\small
	\caption{Evaluating accuracy and fairness performance associated with sensitive attribute ESCS}
	\vspace{-2mm}
	\label{tab:over_escs}
	\resizebox{\linewidth}{!}{
		\begin{tabular}{cc*{12}{c}}
			\toprule
			\multicolumn{2}{c}{\multirow{2}{*}{\textbf{Model}}} & \multicolumn{6}{c}{\textbf{Australia}} & \multicolumn{6}{c}{\textbf{Brazil}} \\
			\cmidrule(lr){3-8} \cmidrule(lr){9-14}
			& & \textbf{EO$\downarrow$} & \text{$\textbf{D}_\textit{disadv}^\textit{under}$} & \textbf{IR$\uparrow$} & \textbf{AUC$\uparrow$} & \textbf{ACC$\uparrow$} & \textbf{DOA$\uparrow$} & \textbf{EO$\downarrow$} & \text{$\textbf{D}_{\textit{disadv}}^{\textit{under}}$}
			& \textbf{IR$\uparrow$} & \textbf{AUC$\uparrow$} & \textbf{ACC$\uparrow$} & \textbf{DOA$\uparrow$} \\
			\midrule
			\multirow{5}{*}{\textbf{IRT}} & Base & 0.0338 & 0.0826 & 0.7353 & 0.7979 & 0.7266 & - & 0.0582 & 0.1407 & 0.5018 & 0.7794 & 0.7269 & - \\
			& Base\rlap{\textsuperscript{†}} & 0.0604 & 0.1473 & 0.7025 & \textbf{0.8080} & \textbf{0.7322} & - & 0.1025 & 0.2510 & 0.4700 & \textbf{0.7958} & \textbf{0.7324} & - \\
			\cmidrule(lr){2-14}
			& Reg & 0.0110 & 0.0270 & \textbf{0.7544} & 0.7961 & 0.7249 & - & 0.0277 & 0.0665 & 0.5301 & 0.7769 & 0.7250 & - \\
			& Adv & 0.0286 & 0.0697 & 0.7449 & 0.7969 & 0.7264 & - & 0.0669 & 0.1609 & 0.4935 & 0.7797 & 0.7268 & - \\
			& \name & \textbf{0.0051} & \textbf{0.0002} & 0.7339 & 0.8022 & 0.7249 & - & \textbf{0.0162} & \textbf{0.0357} & \textbf{0.5760} & 0.7893 & 0.7255 & - \\
			\cmidrule(lr){1-14}
			\multirow{5}{*}{\textbf{MIRT}} & Base & 0.0575 & 0.1408 & 0.7013 & 0.8027 & 0.7299 & - & 0.0913 & 0.2227 & 0.5109 & 0.7836 & 0.7280 & - \\
			& Base\rlap{\textsuperscript{†}} & 0.0645 & 0.1523 & 0.6973 & \textbf{0.8088} & \textbf{0.7339} & - & 0.1251 & 0.3053 & 0.4663 & \textbf{0.7950} & \textbf{0.7316} & - \\
			\cmidrule(lr){2-14} & Reg & 0.0284 & 0.0694 & 0.7279 & 0.8010 & 0.7278 &  - & 0.0512 & 0.1246 & \textbf{0.5539} & 0.7813 & 0.7258 & - \\
			& Adv & 0.0554 & 0.1357 & 0.7009 & 0.8030 & 0.7288 & - & 0.0956 & 0.2335 & 0.5036 & 0.7840 & 0.7283 & - \\
			& \name & \textbf{0.0098} & \textbf{0.0227} & \textbf{0.7520} & 0.7983 & 0.7237 & - & \textbf{0.0279} & \textbf{0.0403} & 0.5248 & 0.7804 & 0.7205 & - \\
			\cmidrule(lr){1-14}
			\multirow{5}{*}{\textbf{NCD}} & Base & 0.0425 & 0.1040 & 0.7183 & 0.7868 & 0.7170 & 0.6248 & 0.0669 & 0.1588 & 0.5220 & 0.7675 & 0.7140 & 0.5972 \\
			& Base\rlap{\textsuperscript{†}} & 0.0857 & 0.2039 & 0.6615 & 0.7911 & 0.7199 & 0.6384 & 0.1274 & 0.3108 & 0.4491 & 0.7718 & 0.7166 & 0.6394 \\
			\cmidrule(lr){2-14}
			& Reg & 0.0331 & 0.0811 & 0.7277 & 0.7863 & 0.7172 & 0.6245 & 0.0522 & 0.1229 & 0.5370 & 0.7669 & 0.7131 & 0.5965 \\
			& Adv & 0.0528 & 0.1292 & 0.6644 & 0.7801 & 0.7111 & 0.5715 & 0.0506 & 0.1234 & 0.5388 & 0.7601 & 0.7112 & 0.5648 \\
			& \name & \textbf{0.0029} & \textbf{0.0010} & \textbf{0.7538} & \textbf{0.7997} & \textbf{0.7234} & \textbf{0.7040} & \textbf{0.0030} & \textbf{0.0028} & \textbf{0.5599} & \textbf{0.7788} & \textbf{0.7209} & \textbf{0.6806} \\
			\cmidrule(lr){1-14}
			\multirow{5}{*}{\textbf{KaNCD}} & Base & 0.0464 & 0.1133 & 0.7113 & 0.8017 & 0.7273 & 0.6584 & 0.0742 & 0.1792 & 0.4877 & 0.7793 & 0.7221 & 0.6046 \\
			& Base\rlap{\textsuperscript{†}} & 0.0770 & 0.1878 & 0.6957 & \textbf{0.8076} & \textbf{0.7310} & 0.6917 & 0.1210 & 0.2963 & 0.5103 & \textbf{0.7910} & \textbf{0.7284} & \textbf{0.6848} \\
			\cmidrule(lr){2-14}
			& Reg & 0.0255 & 0.0622 & 0.7299 & 0.8004 & 0.7260 & 0.6552 & 0.0464 & 0.1115 & 0.5138 & 0.7775 & 0.7207 & 0.6015 \\
			& Adv & 0.0532 & 0.1303 & 0.7075 & 0.8009 & 0.7282 & 0.6615 & 0.0686 & 0.1664 & 0.5388 & 0.7802 & 0.7244 &0.6357 \\
			& \name & \textbf{0.0110} & \textbf{0.0252} & \textbf{0.7484} & 0.8045 & 0.7299 & \textbf{0.7013} & \textbf{0.0363} & \textbf{0.0888} & \textbf{0.5145} & 0.7892 & 0.7267 & 0.6840 \\
			\bottomrule
		\end{tabular}
	}
	\vspace{-4mm}
\end{table*}
Based on the principle of Equalized Odds~\cite{hardt2016equality}, we propose the following evaluation metric to evaluate the fairness performance of models:
\begin{equation}
	\begin{split}
		D^{under}_{disadv}=FNR_{disadv}-FNR_{adv} ,
	\end{split}
	\label{eq:under}
\end{equation}
where $\{FNR_{disadv}$, $FNR_{adv}\}$ denote the False Negative Rates (FNRs) of disadvantaged and advantaged groups. The closer the value of $D^{under}_{disadv}$ is to 0, the better fairness performance the model is.

Meanwhile, for disadvantaged groups, we should also identify the top students as accurately as possible, so that they have opportunities to access higher levels of education. 
Thus, we select the absolute metric F2-score~\cite{f22020unbalanced} to assess the proportion of high-achieving students of disadvantaged groups, which we name as Identified Rate (IR) and formulate as follows:
\begin{equation}
	\begin{split}
		IR = \frac{5\times precision_{disadv}\times recall_{disadv}}{(4\times precision_{disadv})+recall_{disadv}}. \\
	\end{split}
\end{equation}
where $precision_{disadv}$ and $recall_{disadv}$ denote the Precision and Recall of disadvantaged group. 
Note that the larger the value of $IR$ is, the better performance the model has.

\textbf{Implementation Details.} 
As our proposed method is model-agnostic, we apply \name~to four advanced CDMs to show its effectiveness and flexibility. 
Moreover, we compare with several recent approaches, including regularization-based methods and adversarial-based methods: 
\textit{1) Base}: Basic cognitive diagnosis models (i.e., IRT, MIRT, NCD, and KaNCD) that do not consider bias; 
\textit{2) $Base^\dagger$}: Basic Models with Sensitive Attributes;
\textit{3) Reg}: Regularization-based models, by adding Equal opportunity as a regularization to CDMs \cite{hardt2016equality, li2021user};
\textit{4) Adv}: Adversarial learning methods \cite{bose2019compositional} to reduce relevance between sensitive attributes and student representation 

For implementation, we set the learning rate to $0.001$ and batch size to $512$. We apply Adam as the optimization algorithm to update the model parameters. To obtain the best performance, we tune hyper-parameters on validation sets to select the best. 
The balance parameters in Eq.\eqref{eq:training} are set to $1.0, 0.1, 0.5$, and $1.0$, respectively. 

\subsection{Overall Experiments}
Table~\ref{tab:over_escs} and Table~\ref{tab:over_fisced} report overall results under different sensitive attributes. 
We observe all basic CDMs suffered from unfair
outcome issues.
Besides, \textit{$Base^\dagger$} incorporates sensitive attributes into model learning, achieving better diagnosis performance. 
Meanwhile, it also exhibits more severe unfairness. 
All these suggest the urgency to explore fairness-aware learning in CD models. 
Therefore, we evaluate the model performance from the following three aspects: 

For \textit{fairness performance} (i.e., EO and $D_\textit{disadv}^\textit{under}$), we observe that \name~outperforms most baselines, proving its effectiveness. 
Moreover, Adversarial-based methods (Adv) show worse performance than regularization-based (Reg) methods in most cases. 
One possible reason is that the latter directly utilizes sensitive attribute group labels to optimize corresponding metrics. 
In contrast, \name~decouples sensitive attributes and only eliminates the fairness-related sensitive features that should not be used in diagnosis process. 

For the \textit{trade-off between fairness and diagnosis performance}, all debiased baselines perform worse on diagnosis performance than \name, proving that they cannot retain diagnosis-related features from sensitive attributes, which causes a larger decrease in diagnosis accuracy. 
\name~uses the newly designed DP module and multi-factor constraint to retain diagnosis-related features as much as possible, thus outperforming baselines.

For the \textit{Identification capability} (i.e., IR) of high-achieving students from disadvantaged groups, \name~still achieves impressive performance, which proves that \name~can effectively ensure the rights of vulnerable groups. 
Moreover, by considering the results on $D_\textit{disadv}^\textit{under}$ metric, we can conclude that for different CD backbones, \name~can effectively balance fairness and diagnosis performance, demonstrating the flexibility and effectiveness of \name.

\begin{table*}
	\centering
	\small
	\caption{Evaluating accuracy and fairness performance associated with sensitive attribute Father's education level.}
	\vspace{-2mm}
	\label{tab:over_fisced}
	\resizebox{\linewidth}{!}{
		\begin{tabular}{cc*{12}{c}}
			\toprule
			\multicolumn{2}{c}{\multirow{2}{*}{\textbf{Model}}} & \multicolumn{6}{c}{\textbf{Australia}} & \multicolumn{6}{c}{\textbf{Brazil}} \\
			\cmidrule(lr){3-8} \cmidrule(lr){9-14}
			& & \textbf{EO$\downarrow$} & \text{$\textbf{D}_\textit{disadv}^\textit{under}$} & \textbf{IR$\uparrow$} & \textbf{AUC$\uparrow$} & \textbf{ACC$\uparrow$} & \textbf{DOA$\uparrow$} & \textbf{EO$\downarrow$} & \text{$\textbf{D}_{\textit{disadv}}^{\textit{under}}$} & \textbf{IR$\uparrow$} & \textbf{AUC$\uparrow$} & \textbf{ACC$\uparrow$} & \textbf{DOA$\uparrow$} \\
			\midrule
			\multirow{5}{*}{\textbf{IRT}} & Base & 0.0293 & 0.0705 & 0.7346 & 0.7979 & 0.7266 & - & 0.0366 & 0.0896 & 0.5209 & 0.7794 & 0.7269 & - \\
			& Base\rlap{\textsuperscript{†}} & 0.0440 & 0.1069 & 0.6980 & \textbf{0.8087} & \textbf{0.7312} & - & 0.0720 & 0.1723 & 0.5379 & \textbf{0.7959} & \textbf{0.7314} & - \\
			\cmidrule(lr){2-14}
			& Reg & \textbf{0.0132} & 0.0303 & \textbf{0.7515} & 0.7969 & 0.7255 & - & 0.0190 & 0.0465 & 0.5468 & 0.7781 & 0.7262 & - \\
			& Adv & 0.0231 & 0.0546 & 0.7319 & 0.7919 & 0.7200 & - & 0.0314 & 0.0716 & 0.5404 & 0.7752 & 0.7242 & - \\
			& \name & 0.0162 & \textbf{0.0015} & 0.7342 & 0.8034 & 0.7277 & - & \textbf{0.0021} & \textbf{-0.0049} & \textbf{0.5640} & 0.7911 & 0.7274 & - \\
			\cmidrule(lr){1-14}
			\multirow{5}{*}{\textbf{MIRT}} & Base & 0.0437 & 0.1051 & 0.7084 & 0.8027 & 0.7299 & - & 0.0572 & 0.1398 & 0.5472 & 0.7836 & 0.7280 & - \\
			& Base\rlap{\textsuperscript{†}} & 0.0554 & 0.1326 & 0.7195 & \textbf{0.8106} & \textbf{0.7347} & - & 0.0849 & 0.1820 & 0.4931 & \textbf{0.7896} & 0.7279 & - \\
			\cmidrule(lr){2-14} & Reg & \textbf{0.0194} & \textbf{0.0449} & 0.7383 & 0.8025 & 0.7297 &  - & 0.0300 & 0.0735 & 0.5787 & 0.7812 & 0.7272 & - \\
			& Adv & 0.0389 & 0.0947 & 0.7110 & 0.8043 & 0.7316 & - & 0.0528 & 0.1292 & 0.5548 & 0.7821 & \textbf{0.7282} & - \\
			& \name & \textbf{0.0194} & 0.0474 & 0.7095 & 0.8073 & 0.7285 & - & \textbf{0.0247} & \textbf{0.0422} & \textbf{0.5879} & 0.7827 & 0.7214 & - \\
			\cmidrule(lr){1-14}
			\multirow{5}{*}{\textbf{NCD}} & Base & 0.0313 & 0.0747 & 0.7265 & 0.7868 & 0.7170 & 0.6248 & 0.0428 & 0.1042 & 0.5409 & 0.7675 & 0.7140 & 0.5972 \\
			& Base\rlap{\textsuperscript{†}} & 0.0477 & 0.1147 & 0.6981 & \textbf{0.8021} & 0.7263 & 0.6478 & 0.0855 & 0.1745 & \textbf{0.6095} & 0.7785 & 0.7026 & 0.6518 \\
			\cmidrule(lr){2-14}
			& Reg & 0.0293 & 0.0679 & 0.6940 & 0.7834 & 0.7119 & 0.6167 & 0.0324 & 0.0794 & 0.5396 & 0.7689 & 0.7145 & 0.6010 \\
			& Adv & 0.0323 &  0.0789 & 0.6791 & 0.7825 & 0.7130 & 0.5918 & 0.0484 & 0.1177 & 0.5470 & 0.7635 & 0.7150 & 0.5722 \\
			& \name & \textbf{0.0227} & \textbf{-0.0116} & \textbf{0.7362} & 0.8003 & \textbf{0.7276} & \textbf{0.7096} & \textbf{0.0280} & \textbf{0.0465} & 0.5745 & \textbf{0.7879} & \textbf{0.7293} & \textbf{0.6889} \\
			\cmidrule(lr){1-14}
			\multirow{5}{*}{\textbf{KaNCD}} & Base & 0.0370 & 0.0887 & 0.7146 & 0.8017 & 0.7273 & 0.6584 & 0.0433 & 0.1058 & 0.5189 & 0.7793 & 0.7221 & 0.6046 \\
			& Base\rlap{\textsuperscript{†}} & 0.051 & 0.1207 & 0.6938 & \textbf{0.8084} & \textbf{0.7310} & \textbf{0.7183} & 0.0555 & 0.1302 & 0.5434 & 0.7862 & 0.7256 & 0.6731  \\
			\cmidrule(lr){2-14}
			& Reg & 0.0251 & 0.0578 & 0.7364 & 0.8010 & 0.7274 & 0.6642 & \textbf{0.0288} & \textbf{0.0699} & \textbf{0.5826} & 0.7799 & 0.7242 &  0.6347\\
			& Adv & 0.0405 & 0.0972 & 0.7144 & 0.8006 & 0.7278 & 0.6618 & 0.0419 & 0.1020 & 0.5609 & 0.7802 & 0.7239 & 0.6352 \\
			& \name & \textbf{0.0114} & \textbf{0.0275} & \textbf{0.7768} & 0.8066 & 0.7269 & 0.7097 & 0.0340 & 0.0746 & 0.5130 & \textbf{0.7930} & \textbf{0.7278} & \textbf{0.6847} \\
			\bottomrule
		\end{tabular}
	}
	\vspace{-4mm}
\end{table*}
\subsection{Ablation Study}
To verify the effectiveness of each component, we conduct an ablation study with ESCS and report results in Table~\ref{tab:ablation}. 
From the results, when only using $\mathcal{L}_{\text{ce}}$, $\bm{U}_f$ and $\bm{U}_d$ cannot decouple sensitive attributes effectively. Instead, they would introduce more bias, resulting in better diagnosis and worse fairness performances.
When separately introducing $\mathcal{L}_{\text{cls}}$ and $\mathcal{L}_{\text{rev}}$, we observe improvements in fairness performance with minimal impact on accuracy, proving the effectiveness of learned $\bm{U}_f$ and $\bm{U}_d$. 
When using only $\mathcal{L}_{\text{cons}}$, \name~can learn how to remove fairness-aware sensitive features, achieving a substantial improvement in fairness. 
However, since $\bm{U}_f$ still contains some factors affecting fairness, this component can only generate a suboptimal outcome.
Moreover, when removing each component, we observe varying degrees of performance degradation. 
Among all components, removing $\mathcal{L}_{\text{cons}}$ causes the most significant decrease in fairness performance and increase in diagnosis performance. 
This phenomenon not only proves the importance of multi-factor fairness constraint, but also shows \name~will abuse sensitive attributes to improve diagnosis performance. 
Moreover, $\mathcal{L}_{\text{cons}}^*$ denotes the removal of constraints on $\bm{U}_d$ for direct comparison with regularization methods. 
The result proves that the absence of constraints on $\bm{U}_d$ severely impacts fairness performance, aligning its performance comparably with regularization methods.
Furthermore, we observe that $\mathcal{L}_{\text{cls}}$ and $\mathcal{L}_{\text{rev}}$ have a positive impact on fairness performance, removing them will cause a decrease in fairness performance. 
In conclusion, these components are all necessary for the superiority of \name.
\begin{table}
	\centering
	\small
	\caption{Ablation Study of IRT on the Australia Dataset}
	\vspace{-4mm}
	\label{tab:ablation}
	\begin{tabular}{cccccc}
		\toprule
		Conditions & \textbf{EO$\downarrow$} & \text{$\textbf{D}_\textit{disadv}^\textit{under}$} & \textbf{IR$\uparrow$} & \textbf{AUC$\uparrow$} & \textbf{ACC$\uparrow$} \\
		\midrule
		Base & 0.0338 & 0.0826 & 0.7353 & 0.7979 & 0.7266 \\
		\name & \textbf{0.0051} & \textbf{0.0002} & 0.7339 & 0.8022 & 0.7249 \\
		\midrule
		w $\mathcal{L}_{\text{ce}}$ & 0.0545 & 0.1335 & 0.6885 & 0.8089 & 0.7329 \\
		w $\mathcal{L}_{\text{cls}}$ & 0.0503 & 0.1231 & 0.6982 & 0.8088 & 0.7328 \\
		w $\mathcal{L}_{\text{rev}}$ & 0.0525 & 0.1287 & 0.6919 & 0.8090 & 0.7332 \\
		w $\mathcal{L}_{\text{cons}}$ & 0.0088 & 0.0069 & 0.7392 & 0.8016 & 0.7250 \\
		\midrule
		w/o $\mathcal{L}_{\text{cls}}$ & 0.0137 & 0.0318 & 0.7206 & 0.8057 & 0.7279 \\
		w/o $\mathcal{L}_{\text{rev}}$ & 0.0112 & -0.0057 & 0.7277 & 0.8022 & 0.7258 \\
		w/o $\mathcal{L}_{\text{cons}}$ & 0.0609 & 0.1493 & 0.7127 & \textbf{0.8092} & \textbf{0.7337} \\
		w/o $\mathcal{L}_{\text{cons}}^*$ & 0.0132 & -0.0322 & \textbf{0.7565} & 0.8021 & 0.7257 \\
		\bottomrule
	\end{tabular}
	\vspace{-4mm}
\end{table}

\subsection{Parameter Sensitive Test}
\begin{figure}[htbp]
	\vspace{-2mm}
	\begin{center}
		\includegraphics[width=0.41\textwidth]{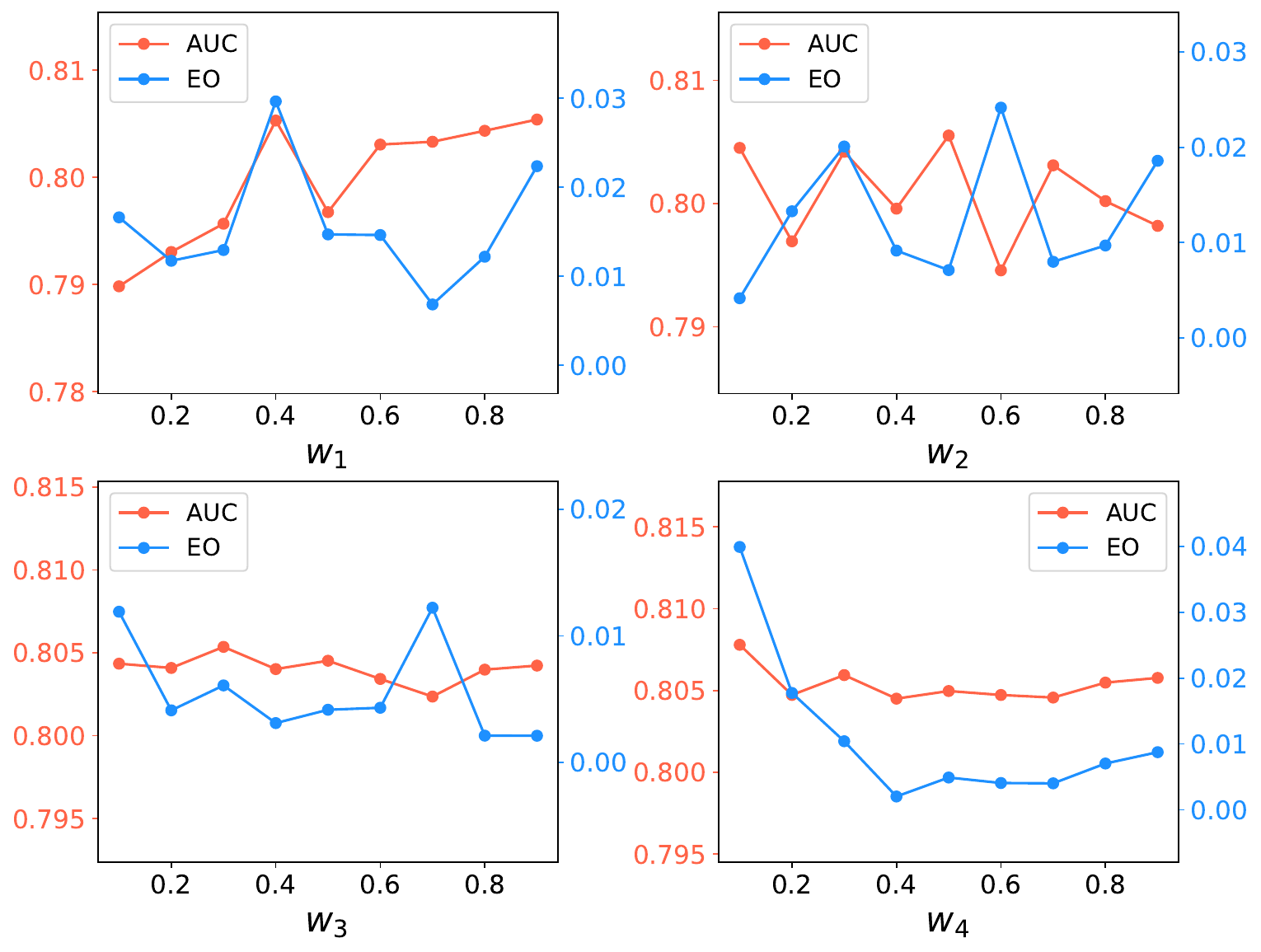}
	\end{center}
	\vspace{-3mm}
	\caption{\name~ based on IRT with varying $w$}
	\label{f:w}
	\vspace{-6mm}
\end{figure}

To conduct a deeper analysis on \name, we also conduct parameter sensitive test on the weights $w_1$, $w_2$, $w_3$, $w_4$ in Eq.(\ref{eq:training}), whose results are summarized in Figure~\ref{f:w}. 
According to the results, we can observe that with the increase of $w_4$, the diagnosis performance of \name~decreases slightly while the fairness performance increases rapidly. 
This phenomenon is consistent with the results in Table~\ref{tab:ablation}, proving the importance of $\mathcal{L}_{\text{cons}}$. 
Moreover, with the weight $w_1$ increasing, the diagnosis performance of \name~increases while the fairness performance decreases rapidly and is unstable. 
This is intuitive since a large $w_1$ will focus more on student data and impose more biased information. 
With the increase of $w_2$, both diagnosis performance and fairness performance fluctuate considerably. 
We speculate one possible reason is that $\mathcal{L}_{\text{cls}}$ is relatively simple and its impact on \name~is relatively small. Thus, a large weight will lead \name~unable to learn useful information.  
For $w_3$, a larger value will improve the fairness performance while retaining the diagnosis performance, proving the effectiveness of $\mathcal{L}_{\text{rev}}$.

\begin{figure}[htbp]
	\vspace{-2mm}
	\begin{center}
		\includegraphics[width=0.43\textwidth]{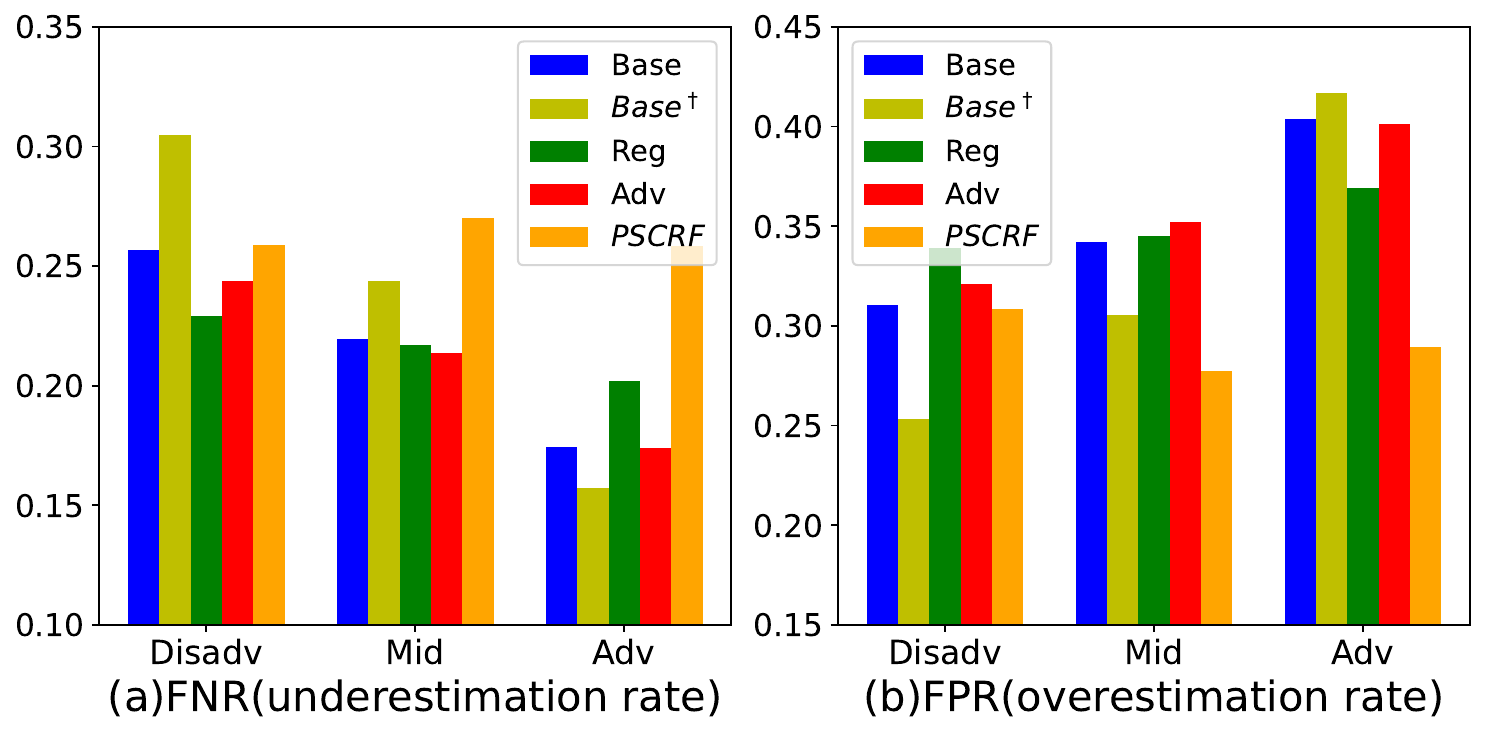}
	\end{center}
	\vspace{-3mm}
	\caption{Visualization of the prediction distributions.}
	\label{f:case}
	\vspace{-6mm}
\end{figure}
\subsection{Case Study}
To better illustrate the effectiveness of \name, we visualize the prediction distributions of different models using FNR (underestimate rate) and FPR (overestimate rate). 
As shown in Figure~\ref{f:case}, we initially observe a conspicuous prediction bias in the $Base$ model, characterized by underestimation for the disadvantaged group and overestimation for the advantaged group, displaying a distinct stepwise distribution between different groups. 
When introducing sensitive attributes, this unfairness is exacerbated (e.g., $Base^\dagger$ model). 
$Reg$ and $Adv$ methods can alleviate this unfairness to some extent. However, their prediction distributions still exhibit a stepwise pattern. 
As a comparison, \name~shows nearly consistent levels of underestimation and overestimation across different groups, achieving the lowest overestimation rates among the various models. 
A comparison with $Base^\dagger$ reveals that \name~has effectively mitigated the unfairness imposed on the disadvantaged group by a single sensitive attribute.
The only shortfall is that our approach does not reduce the underestimation rates.
We speculate that there might exist other sensitive attributes that are not considered.

\section{Conclusion}
\label{s:conclusion}
In this paper, we argued that current CD models concentrated more on the full exploitation of student-exercise interaction data, ignoring the potential risk of abuse of sensitive attributes. 
Moreover, we argued that student sensitive attributes could also provide useful information, so directly eliminating them was not optimal. 
To achieve fairness-aware CD learning, we proposed to incorporate casual inference and designed a novel \name. 
By leveraging a newly designed attribute-oriented predictor to deal with the information from sensitive attributes. 
\name~decoupled sensitive attributes into 
fairness-related sensitive features that should not be used in the diagnosis process, and diagnosis-related features that should be used to enrich the student proficiency level modeling. 
Thus, \name~could achieve impressive fairness performance while retaining the diagnosis performance. 
Moreover, we designed a multi-factor fairness constraint to ensure the fairness performance and diagnosis performance simultaneously.  
Finally, we conducted extensive experiments over real-world datasets and multiple advanced CD models to demonstrate the effectiveness of \name.

\section{Acknowledgments}
This work was supported in part by grants from the National Science and Technology Major Project (Grant No. 2021ZD0111802), Joint Funds of the National Natural Science Foundation of China (Grant No. U22A2094), and the National Natural Science Foundation of China (Grant No. 62376086, U23B2031, 721881011).

\bibliographystyle{ACM-Reference-Format}
\balance
\bibliography{7-reference}

\newpage
\appendix
\section{Appendix}
\label{s:appendix}

\begin{table*}[htbp]
	\centering
	\caption{The statistics of useful but not sensitive attributes associated with ESCS}
	\vspace{-4mm}
	\label{tab:useful1}
	\begin{tabular}{c|c|c|c}
		\toprule
		id & name & correlation & category \\
		\midrule
		ST013Q01TA & How many books are there in your home? & 0.416 & 0: 0-100 books, 1: More than 100 books \\
		ST012Q07NA & Tablet computers & 0.402 & 0: Zero or one 1: More than one \\
		ST011Q06TA & A link to the Internet & 0.308 & 0: Yes 1: No \\
		STo11Q04TA & A computer you can use for school work & 0.301 & 0: Yes 1: No \\
		ST012Q08NA & E-book readers & 0.266 & 0: NaN 1: More than one \\
		\bottomrule
	\end{tabular}
	\vspace{0mm} 
\end{table*}
\begin{table*}[htbp]
	\centering
	\caption{The statistics of useful but not sensitive attributes associated with Father's education level}
	\vspace{-4mm}
	\label{tab:useful2}
	\begin{tabular}{c|c|c|c}
		\toprule
		id & name & correlation & category \\
		\midrule
		ST013Q01TA & How many books are there in your home? & 0.286 & 0: 0-100 books, 1: More than 100 books \\
		ST012Q07NA & Tablet computers & 0.166 & 0: Zero or one 1: More than one \\
		ST011Q09TA & Works of art (e.g. paintings) & 0.165 & 0: Yes 1: No \\
		ST011Q16NA & Books on art, music or design & 0.146 & 0: Yes 1: No \\
		ST011Q07TA & Classic literature (e.g. <Shakespeare>) & 0.142 & 0: Yes 1: No \\
		\bottomrule
	\end{tabular}
	\vspace{0mm}
\end{table*}
\subsection{The selection of Educational Conetxt}
	We present the most relevant useful attributes associated with different sensitive attributes in Table \ref{tab:useful1} and Table \ref{tab:useful2}.
	We employ the Pearson correlation coefficient to calculate the correlation, and select the Top-k attributes as the educational context used in Eq.(\ref{eq:non-sensitive}) in Section~\ref{s:constraint}:
	\begin{equation}
		\rho_{X,Y} = \frac{\sum_{i=1}^{n}(X_i - \bar{X})(Y_i - \bar{Y})}{\sqrt{\sum_{i=1}^{n}(X_i - \bar{X})^2}\sqrt{\sum_{i=1}^{n}(Y_i - \bar{Y})^2}}
	\end{equation}
	\subsection{The Performance of \name~ on Graph-Based RCD Model}
	Since our \name~ focuses more on fairness-aware cognitive diagnosis, we select fundamental and general CD models in the main text. To further demonstrate the generality of \name~, we applied PSCRF to the graph-based model RCD\cite{gao2021rcd} and report the results on Australia dataset regarding the sensitive attribute ESCS in Table \ref{tab:rcd_res}. It can be observed that \name~ achieves similarly good performance on the RCD model as well.
	\subsection{The relevant information about the learnable parameters $\alpha$ and $\beta$.}
	We want to know the values of $\alpha$ and $\beta$ after they have been learned, so we computed the mean and variance of their values after training on the Australia dataset using IRT model, as shown in Table \ref{tab:learned_param}. From the results, we can conclude that our proposed \name~ realizes user-specific debiasing according to the input user's situation.
	

\begin{table*}[htbp]
	\centering
	\caption{The Performance of \name~on Graph-Based RCD Model}
	\begin{tabular}{ccccccc}
		\toprule
		\textbf{Method} & \textbf{EO$\downarrow$} &  \text{$\textbf{D}_\textit{disadv}^\textit{under}$} & \textbf{IR$\uparrow$} & \textbf{AUC$\uparrow$} & \textbf{ACC$\uparrow$} & \textbf{DOA$\uparrow$}  \\
		\midrule
		Base & 0.0394 & 0.0965 & 0.7132 & 0.7971 & 0.7265 & 0.6209 \\
		Base\rlap{\textsuperscript{†}} & 0.0622 & 0.1523 & 0.7000 & 0.8005 & 0.7269 & 0.6744 \\
		\name~ & 0.0032 & 0.0004 & 0.7433 & 0.8009 & 0.7265 & 0.6327 \\
		\bottomrule
	\end{tabular}
	
	\label{tab:rcd_res}
\end{table*}	

\begin{table}[t]
	\centering
	\caption{The statistical data for alpha and beta.}
	\begin{tabular}{ccc}
		\toprule
		\textbf{param} & \textbf{mean} &  \textbf{variance} \\
		\midrule
		$\alpha$ & 0.2043 & 0.0826 \\
		$\beta$ & 0.4041 & 0.1304 \\
		\bottomrule
	\end{tabular}
	
	\label{tab:learned_param}
\end{table}

\end{document}